\documentclass{article}
\usepackage{ijcai26}
\usepackage{tcolorbox}
\usepackage{amssymb}

\usepackage{times}
\usepackage{soul}
\usepackage{url}
\usepackage[hidelinks]{hyperref}
\usepackage[utf8]{inputenc}
\usepackage[small]{caption}
\usepackage{graphicx}
\usepackage{amsmath}
\usepackage{amsthm}
\usepackage{booktabs}
\usepackage{algorithm}
\usepackage{algorithmic}
\usepackage[switch]{lineno}
\usepackage{xcolor}

\title{Beyond Isolated Investor: Predicting Startup Success via Roleplay-Based
Collective Agents}

\author{
Zhongyang Liu$^1$
\and
Haoyu Pei$^1$
\and
Xiangyi Xiao$^1$
\and
Xiaocong Du$^1$
\and
Yihui Li$^1$
\and
Suting Hong$^2$
\and
Kunpeng Zhang$^3$
\and
Haipeng Zhang$^1$\thanks{Corresponding author} 
\affiliations
$^1$ShanghaiTech University\\
$^2$Xi'an Jiaotong-Liverpool University\\
$^3$University of Maryland\\
\emails
\{liuzhy12024, peihy2024, xiaoxy, duxc2023, liyh, zhanghp\}@shanghaitech.edu.cn,
sutinghong@gmail.com,
kpzhang@umd.edu
}

\begin{document}

\maketitle
\begin{abstract}
Due to the high value and high failure rates of startups, predicting their success is a critical challenge. Existing approaches typically model startup success from a single decision-maker's perspective, overlooking the collective dynamics that dominate real-world venture capital (VC) decision-making. We propose \textbf{SimVC-CAS}, a collective agent system that simulates VC decisions as a multi-agent interaction process. By designing role-playing agents and a GNN-based supervised interaction module, we reformulate startup financing prediction as a group decision-making task, capturing both enterprise fundamentals and investor network dynamics. Each agent represents an investor with distinct traits and preferences, enabling heterogeneous evaluations and realistic information exchange over a graph-structured co-investment network.
Using both proprietary and public VC data with strict anti-leakage controls, we show that SimVC-CAS significantly improves predictive performance, achieving approximately 25\% relative improvement in average precision@10, while exhibiting consistency with real investor decisions. The interaction mechanism is particularly effective for network-central startups, confirming the importance of network in VC decision-making. Analysis of agents' reasoning for decision changes further reveals how network environment influence decision quality, demonstrating the system's interpretability. Our approach may generalize to broader group decision-making scenarios.
Our code is available~\footnote{https://anonymous.4open.science/r/SimVC-CAS-888C}.
\end{abstract}

\section{Introduction}
Startups contributed 60\% of global new jobs (2010-2020)~\cite{european2019missing} 
yet face a 90\% failure rate~\cite{kerr2014entrepreneurship}. This significant contrast has made the prediction of early-stage startup success (e.g., financing outcomes and IPO potential) a central problem in interdisciplinary research~\cite{zhang2021scalable,10.1145/3763001}. Among various success indicators, early-stage financing plays a critical role, serving as the ``lifeline" for startup survival and growth~\cite{cassar2004financing}. Access to capital not only provides essential financial resources but also shapes long-term strategic direction. Consequently, predicting whether a startup can successfully secure financing has emerged as a core task in startup success prediction~\cite{zhang2021scalable,dellermann2021finding}.

In venture capital (VC), investors typically base their decisions on comprehensive evaluations that consider factors such as team composition, business model, and market potential~\cite{miloud2012startup,kim2022venture}. Following this logic, prior automated methods have primarily modeled financing outcomes from the perspective of individual decision-making by extracting firm-level features using various technical paradigms. Representative lines of work include: (i) traditional machine learning on engineered features~\cite{krishna2016predicting,unal2019machine}; (ii) graph neural network (GNN)-based modeling of investment networks~\cite {zhang2021scalable,10.1145/3763001}; and (iii) pre-trained language models for textual analysis~\cite{maarouf2025fused}. \textbf{Despite methodological diversity, these approaches share a common assumption: they aim to simulate a single, idealized investor.}

\begin{figure}[t]
    \centering
    \includegraphics[width=0.7\linewidth]{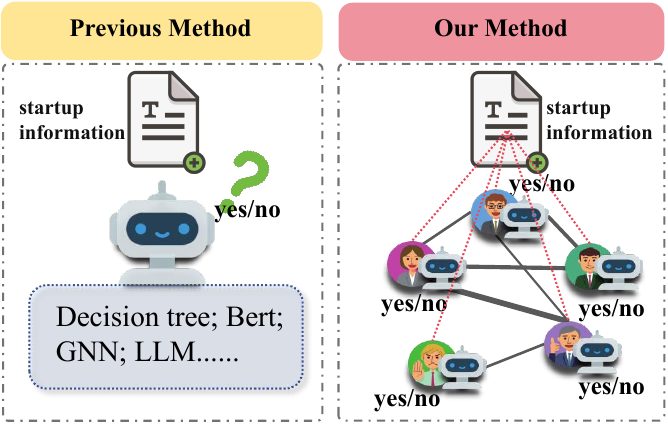}
    \vspace{-5pt}
    \caption{Unlike prior single perspective approaches, our method jointly models the collective behaviors of startups and investors by incorporating  multiple decision-makers.}
    \label{fig:Comparison of methods}
    \vspace{-12pt}
\end{figure}

In reality, investor decisions directly shape startup financing outcomes and subsequent development trajectories. The absence of investment can trigger immediate survival crises, while investors' heterogeneous resources, expertise, and networks can profoundly influence long-term growth paths~\cite{diamond2001banks,miloud2012startup}. \textbf{Importantly, real-world VC evaluations are rarely conducted in isolation. Instead, startups are assessed by multiple potential investors embedded within co-investment networks, where information exchange and peer influence play a central role~\cite{han2013social,liu2017large,goldstein2025information}.} Our data confirm this dynamic: the average graph distance among investors who eventually co-invest in the same startup is only 2.16, compared to 4.23 across the entire VC network. As a result, financing outcomes often emerge from the collective judgment of an investor group, and startups of similar quality may experience markedly different outcomes depending on the composition of their investor collective. This ``investor collective effect" is not captured by existing single-decision-maker models. Given the inherent multi-agent nature of real-world VC decision-making, a multi-agent framework is a natural and necessary modeling choice.


However, existing multi-agent systems fail to capture two defining 
characteristics of VC forecasting: \textbf{(i)} the \textbf{collective 
nature} of syndicated decision-making and \textbf{(ii)} the \textbf{complex 
relational dynamics} among co-investors.
\textbf{First}, most frameworks rely on agent specialization to simulate 
compartmentalized reasoning across information dimensions—assigning 
individual agents to analyze different dimensions of information~\cite{yu2024fincon,zhang2024multimodal,wang2025ssffinvestigatingllmpredictive,pei2026gainingpathsinvestmentsuccess}. 
While effective for simulating the internal reasoning process of a single 
investor, this paradigm fails to capture the diversity and interaction of 
multiple independent investors.
\textbf{Second}, existing agent interaction mechanisms are largely 
heuristic-based~\cite{wang-etal-2025-beyond,qian2025scaling,li2025political} 
and cannot model the nuanced, asymmetric relationships observed in real 
co-investment behavior—relationships shaped by network topology, investor 
heterogeneity, and startup-specific context.

To address these challenges, we propose \textbf{SimVC-CAS}: a collective agent system for simulating real-world venture capital decision-making. The core idea is to reframe financing prediction as the outcome of a simulated collective decision process carried out by a network of investors, each modeled as an autonomous agent. SimVC-CAS jointly captures enterprise fundamentals and multi-investor decision dynamics. Given a target startup, the  system simulates (i) heterogeneous individual evaluations by distinct investor agents and (ii) collective decision updates shaped by a co-investment network. The resulting group-level verdict forms the basis for financing prediction. As shown in Figure~\ref{fig:Comparison of methods}, this framework enables joint modeling of both company-level attributes and investor-collective behaviors, offering a more realistic and interpretable approach to VC forecasting. 

Specifically, SimVC-CAS consists of three modules: \textbf{1. Startup Panoramic Portrait.} This module integrates diverse startup data - including  basic information, founding team backgrounds, and financing history - to construct a comprehensive  profile that serves as shared context for all investor agents. \textbf{2. Heterogeneous Investor Portraits.} We construct the potential investor pool from the startup's historical investors and their past co-investment partners. Empirically, 77.35\% of investors in new financing rounds originate from this group in our dataset. Each investor agent is assigned a distinct profile based on background information, experience, and historical investment behavior. Leveraging the role-playing capabilities of large language models (LLMs)~\cite{shanahan2023role,wang2024rolellm}, agents interpret startup information through personalized lenses, capturing heterogeneous preferences,  cognitive biases, and decision patterns. \textbf{3. Collective Agent Interaction Modeling.} Investor interactions depend not only on their own attributes and network structure, but also vary with the characteristics of the evaluated startup, forming startup-centric interaction modes~\cite{elfring2007networking,lee2015modes}. To model these dynamics, we design a graph attention network with virtual nodes (VGAT), where a virtual node represents the target startup and serves as an information hub connected to all investor agents. Investor nodes are linked via edges derived from historical co-investment relationships. Through VGAT, SimVC-CAS captures multi-dimensional interaction patterns that balance investor heterogeneity, startup characteristics, and graph topology, thereby approximating realistic group deliberation processes. 

Our contributions are summarized as follows:
\begin{itemize}
    \item We introduce a simulation-based paradigm that reframes startup 
    financing prediction as a collective decision-making process by 
    integrating LLM role-playing with multi-agent modeling.
    
    \item We propose \textbf{SimVC-CAS}, which jointly models company 
    fundamentals and investor dynamics through scalable profiles and 
    graph-based interactions, achieving significant performance 
    improvements on proprietary and public datasets, with behavioral 
    alignment with real investors.

\item We demonstrate that the interaction mechanism is particularly 
    effective for network-central startups, confirming the importance of 
    network structure in VC decision-making. Analysis of agents' reasoning 
    for decision changes further reveals how network environment influence 
    decision quality, demonstrating the system's interpretability. Beyond 
    VC, our design may generalize to other group decision-making problems.
    
\end{itemize}


\section{Related Work}

\subsection{Startup Success Prediction}

Predicting early-stage startup success is a critical task. Early research identified key predictive factors ~\cite{song2008success,sevilla2022success} which informed manual feature engineering efforts. Initial methods used commercial datasets such as PitchBook, Crunchbase to design features based on domain knowledge and heuristics for traditional machine learning models ~\cite{unal2019machine,bargagli2021supervised}. As large-scale datasets become available, the field evolved toward three main approaches: (1) graph neural networks (GNNs) ~\cite{zhang2021scalable,10.1145/3763001}, which mine relational structures for investment patterns; (2) pre-trained language models (LMs) ~\cite{maarouf2025fused}, which extract semantic features from company descriptions and filings; and (3) multi-agent systems ~\cite{wang2025ssffinvestigatingllmpredictive,griffin2025randomruleforestrrf,pei2026gainingpathsinvestmentsuccess}, which assign different agents to analyze complementary aspects of a company's objective information, aiming to address the interpretability and scalability limitations.

Despite these advances, existing models all focus on single-investor decision frameworks, failing to capture the multi-perspective evaluation and group dynamics typical of the real world. Even when employing multi-agent approaches, they merely use multiple agents to analyze different dimensional information~\cite{wang2025ssffinvestigatingllmpredictive,griffin2025randomruleforestrrf}, remaining within a single decision-maker perspective. 

\subsection{LLM Agents in Decision Simulation}

Large Language Models have shown significant potential in simulating human intelligence and are increasingly used as a foundational component in autonomous agent systems \cite{luo2025large,he2025llm,zhang2025scientific}. In decision simulation, LLM agents have been deployed to model user search behavior and predict click-through rates \cite{zhang2024usimagent}, forecast legislative bill outcomes based on policymaker behavior \cite{li2025political}, and replicate group voting patterns from demographic data \cite{argyle2023out}.

However, the application of LLMs to simulate investor decision-making in VC remains largely underexplored. Current approaches are insufficient for modeling the heterogeneous, relational dynamics inherent in VC investment networks. To address this gap, we propose a graph neural network-based multi-agent interaction framework for simulating VC investment decisions. Beyond VC, this framework offers a generalizable foundation for decision modeling in other complex, heterogeneous network environments. 

\section{Task Definition}

This study focuses on predicting the success of early-stage startups. Following prior work \cite{zhang2021scalable}, we define early-stage startups as those have completed their first round of formal financing (seed or angel round), but have not yet raised subsequent rounds (e.g., Series A). These companies enter the venture capital ecosystem for the first time through initial investors. While success is often measured by whether a startup secures Series A funding \cite{zhang2021scalable,dellermann2021finding}, prior studies have used different observation windows, potentially introducing time biases. To alleviate this, we adopt a consistent one-year observation window, which aligns with stage-based evaluation practices and helps control for external environmental factors such as market cycles and macroeconomic shocks \cite{boocock1997evaluation}. Accordingly, the core task of this study is defined as: \textbf{predicting whether a startup secures subsequent funding within one year of its initial financing.}

\begin{figure*}[!h]
    \centering
    \includegraphics[width=0.7\linewidth]{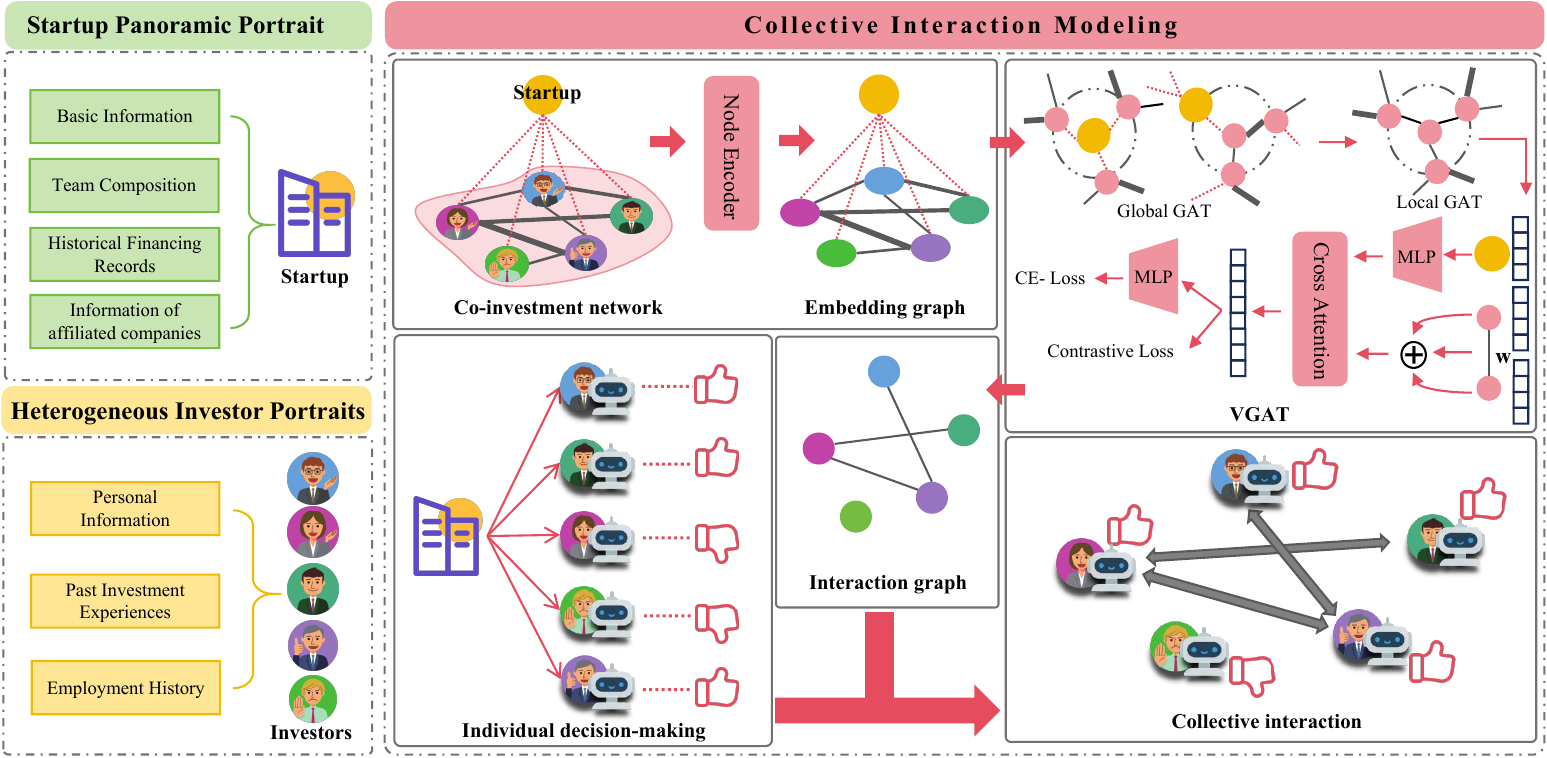}
    \caption{The overall framework of our proposed method. }
    \label{fig:method}
    \vspace{-10pt}
\end{figure*}

\section{Method}
This section details the core framework of our proposed method, shown in Figure  \ref{fig:method}. We begin by constructing a comprehensive profile for each target startup, integrating basic information, team composition, and historical financing records. This profile serves as the fundamental input for investment evaluation. Meanwhile, we generate heterogeneous profiles for potential investors, encompassing personal information, investment experience, and professional background. To simulate investor decision-making, we employ an LLM-based agent  to model the investment preferences of individual investors through role-playing. Building on this, we introduce a collective agent interaction system: for each startup, we construct a co-investment network linking its potential investors based on historical co-investment relationships. In addition, we designed a GAT with virtual nodes (VGAT) to capture the investor interaction patterns centered on startups in this network, enabling us to approximate the collective decision-making dynamics in the real world. This collective modeling enhances the predictive accuracy of startup success.

\subsection{Startup Panoramic Portrait}


To predict success, we construct a panoramic profile capturing the target startup’s foundational characteristics as a holistic input for subsequent agents.

\noindent\textbf{Basic Information:} Founding date, industry, description, 
location, products, keywords, and other core attributes. In classic financial theory, these fundamental characteristics serve as critical signals for screening potential investments and reducing information asymmetry between entrepreneurs and investors~\cite{kaplan2004characteristics,spence1973job}.

\noindent\textbf{Team Composition:} Background  on key team members (gender, education history, prior employment records, personal investment experience, roles), reflecting the ``bet on the jockey'' hypothesis in VC, which posits that human capital is often a more significant predictor of success than the business idea itself~\cite{kaplan2009should,gompers2020how}.

\noindent\textbf{Historical Financing Records:} This section summarizes the startup's past funding rounds, including the amounts raised and investor backgrounds. Reflecting the mechanism of capital staging used to monitor progress and mitigate agency costs~\cite{gompers1995optimal,sahlman1990structure}, these records provide essential cues regarding the firm's valuation trajectory and external validation.

\noindent\textbf{Affiliated Companies:} In VC networks, startups sharing key individuals (e.g., co-investors) with the target often provide valuable context. We define such entities as affiliated companies and incorporate their information into the target's panoramic profile. These shared links facilitate information spillovers and social endorsements, allowing affiliated firms' attributes to serve as high-fidelity signals of the target's potential~\cite{stuart1999interorganizational,hochberg2007whom}.

\subsection{Heterogeneous Investor Portraits}
For each target startup, we construct a real-time, heterogeneous profile of its potential investors. The candidate investor set is constructed based on the following rule: \textbf{We randomly sample $k$ individuals from the union of the startup's historical investors and their past co-investment partners.} Empirically, 77.35\% of investors in new financing rounds originate from this group in our dataset. 
Based on this candidate set, we generate detailed investor profiles and simulate their behavior through a role-playing agent architecture. Each agent represents a distinct investor persona, offering differentiated analysis and decision-making support. This profiling system is highly extensible and can integrate data from multiple sources, such as demographic information, professional experience, and past investment records. Each investor profile includes the following core components:

\noindent\textbf{Personal Information} 
This contains information such as the investor's name, gender and educational background.

\noindent\textbf{Past Investment Experience}
This section details the investor's historical funding activity and is further divided into: (i) early-stage investments - focused on seed and Series A rounds; these are the primary reference points for our model. (ii) Other investments - covers mid-to-late-stage or non-startup investments, serving as supplementary context. All investment records are paired with structured summaries of the invested startups to provide a comprehensive view of the investor's track record.

\noindent\textbf{Employment History}
This details the investors' professional background, including previous roles and affiliations.

\subsection{Collective Agent Interaction Modeling}
To better approximate real-world investor decision-making dynamics, we design a collective agent interaction framework supervised by a graph neural network. In this system, each investor agent first performs an individual evaluation of the target startup based on their profile and outputs an initial investment decision. Next, the historical co-investment network is used (along with VGAT) to determine the interaction structure among these investor agents. Finally, the agents interact according to the learned topology, updating their beliefs and assessments based on peer influence. This dynamic group reasoning process allows agents to revise their initial decisions, resulting in a final, aggregated investment decision that reflects both individual judgment and collective insight.

\vspace{-3pt}
\subsubsection{VGAT}
To model investor interaction patterns shaped jointly by the characteristics of the target startup, the attributes of individual investors, and the topology of the co-investment network, we design the Virtual Node Graph Attention Network (VGAT). In this architecture, the target company is introduced as a virtual node, serving as a global query to attend to relevant investor interactions.

Formally, let $\mathcal{G} = (\mathcal{V}, \mathcal{E}, \omega)$ be an undirected weighted graph. The node set is defined as $\mathcal{V} = \mathcal{V}_R \cup \{v_{target}\}$, where $\mathcal{V}_R = \{v_1,\dots,v_n\}$ represents real investor nodes and $v_{target} \notin \mathcal{V}_R$ denotes the virtual node (i.e., the target company). The edge set consists of two parts: $\mathcal{E} = \mathcal{E}_R \cup \mathcal{E}_V$, where $\mathcal{E}_R \subseteq \mathcal{V}_R \times \mathcal{V}_R$ represents real investor connections, and $\mathcal{E}_V = \{ (v_{target}, v_i) \mid \forall v_i \in \mathcal{V}_R \}$ connects the virtual node to all real nodes. A weight function $\omega: \mathcal{E} \to \mathbb{R}^+$ assigns positive weights to all edges.

The VGAT model processes $\mathcal{G}$ through three sequential components. First, a global Graph Attention (GAT) layer updates embeddings for all nodes:
\begin{equation}
\{\boldsymbol{h}_v^{(1)}\}_{v\in\mathcal{V}} = \text{GAT}(\mathcal{G}).
\end{equation}

\noindent Second, a local GAT layer refines the embeddings specifically for real investor nodes:
\begin{equation}
\{\boldsymbol{h}_v^{(2)}\}_{v\in\mathcal{V}_R} = \text{GAT}(\mathcal{G}[\mathcal{V}_R]).
\end{equation}

\noindent Concurrently, the virtual node embedding is transformed via an MLP to serve as the global query vector:
\begin{equation}
\boldsymbol{h}_{v_{target}}^{*} = \text{MLP}(\boldsymbol{h}_{v_{target}}^{(1)}).
\end{equation}

\noindent To determine the importance of each co-investment relationship, we compute an attention score for each real edge $e_{ij} = (v_i,v_j) \in \mathcal{E}_R$. The cross-attention mechanism uses the virtual node as the query and the edges as keys. The normalized attention coefficient $\alpha_{ij}$ is computed via a Softmax operation over all edges in the batch:
\begin{equation}
\alpha_{ij} = \text{Softmax}_{e_{ij}\in\mathcal{E}_R}\left( \frac{ (\boldsymbol{h}_{v_{target}}^{*}\boldsymbol{W}_Q) (\boldsymbol{m}_{ij}\boldsymbol{W}_K)^\top }{\sqrt{d_k}} \right),
\end{equation}
\noindent where $d_k$ is the scaling dimension. The final edge embedding $\boldsymbol{e}_{ij}$ is then obtained by weighting the value projection:
\begin{equation}
\boldsymbol{e}_{ij} = \alpha_{ij} (\boldsymbol{m}_{ij}\boldsymbol{W}_V),
\end{equation}
\noindent where the base edge representation $\boldsymbol{m}_{ij}$ concatenates the node features and edge weight:
\begin{equation}
\boldsymbol{m}_{ij} = \left[ \boldsymbol{h}_{v_i}^{(2)} \parallel \boldsymbol{h}_{v_j}^{(2)} \parallel \omega_{ij} \right].
\end{equation}
Here, $\boldsymbol{W}_Q, \boldsymbol{W}_K,$ and $\boldsymbol{W}_V$ are learnable projection matrices.

We formulate the interaction prediction as a \textbf{binary classification task}. Since direct data on investor communications is private and unobservable, we employ \textbf{future co-investment relationships} as a proxy for effective interaction. The rationale is that if two investors co-invest in the same startup within a specific future window, they likely engaged in meaningful interaction or shared similar decision logic to mitigate selection uncertainty~\cite{lerner1994syndication,hochberg2007whom}. Thus, the goal is to predict the existence of a valid future co-investment link (Class 1) versus a non-existent one (Class 0).


The model is optimized jointly using cross-entropy and contrastive losses. The Cross-Entropy Loss is defined as:
\begin{equation}
\mathcal{L}_{\text{CE}} = -\frac{1}{|\mathcal{E}_R|} \sum_{e_{ij}\in\mathcal{E}_R} \sum_{c=1}^{N_C} y_{ij}^c \log \left( \hat{y}_{ij}^c \right),
\end{equation}
where $N_C=2$ is the number of classes. $y_{ij}^c$ indicates the ground truth label derived from future co-investment data, and $\hat{y}_{ij}^c$ is the predicted probability derived from $\boldsymbol{e}_{ij}$.

The Contrastive Loss enhances the distinctiveness of edge embeddings:
\begin{equation}
\mathcal{L}_{\text{Contrast}} = -\frac{1}{|\mathcal{P}|} \sum_{(e_{ij},e_{kl})\in\mathcal{P}} \log \frac{\exp\left( \frac{\boldsymbol{e}_{ij}^\top \boldsymbol{e}_{kl}}{\tau \|\boldsymbol{e}_{ij}\|\|\boldsymbol{e}_{kl}\|} \right)}{\sum_{e_{mn}\in\mathcal{N}_{ij}} \exp\left( \frac{\boldsymbol{e}_{ij}^\top \boldsymbol{e}_{mn}}{\tau \|\boldsymbol{e}_{ij}\|\|\boldsymbol{e}_{mn}\|} \right)},
\end{equation}
where $\mathcal{P}$ is the set of positive pairs, $\mathcal{N}_{ij}$ represents negative samples, and $\tau$ is the temperature parameter. The final loss is:
\begin{equation}
\mathcal{L} = \mathcal{L}_{\text{CE}} + \lambda \mathcal{L}_{\text{Contrast}},
\end{equation}
where $\lambda$ is a hyperparameter balancing the two terms.

\noindent\textbf{Training and Validation:} We strictly adhered to the temporal split strategy of the main experiment (see Sec.\ref{sec:Datasets}) to train and evaluate VGAT: data prior to September 2021 was utilized for training and validation, while subsequent data was reserved for testing. As shown in Table~\ref{tab:VGAT_performance}, VGAT significantly outperforms the GAT baseline (F1: 79.31\% vs. 73.47\%). This confirms that explicitly modeling the startup as a global query node enables the network to more accurately discern context-specific investor interaction patterns. \textit{(Please refer to the \textbf{Appendix 1.2} for VGAT training data construction methods and full training details.)}

\vspace{-5pt}
\begin{table}[h]
\small
\centering
\setlength{\tabcolsep}{6pt}
\caption{Performance comparison on the link prediction task}
\vspace{-8pt}
\label{tab:VGAT_performance}
\begin{tabular}{lccc}
\toprule
Method & Precision & Recall & F1 \\
\midrule
GAT & 69.62 & 77.76 & 73.47 \\
\textbf{VGAT} & \textbf{76.18} & \textbf{82.70} & \textbf{79.31} \\
\bottomrule
\end{tabular}
\vspace{-12pt} 
\end{table}

\subsubsection{Investor Agent Decision-Making}
Given a set of $k$ investor agents $\{A_1, A_2, \dots, A_k\}$, we construct a heterogeneous profile $P_i$ for each investor $A_i$. Through role-playing, conditioned on the startup profile $\mathcal{S}$, each agent generates an initial independent decision $D_i^{(0)}$ using a frozen LLM:
\begin{equation}
D_i^{(0)} = \text{LLM}(\mathcal{S}|P_i)
\end{equation}

We construct a co-investment network $\mathcal{G}$ where nodes $\mathcal{V} = \{A_1, \dots, A_k\} \cup \{v_{target}\}$ represent investors and the target startup. Edges encode historical co-investments and connections to the target. Using a pre-trained \footnote{\url{https://huggingface.co/jinaai/jina-colbert-v2.} We selected this model for its superior long-context processing capabilities } encoder, we embed investor profiles and the startup profile:

\begin{itemize}
  \item Investor embedding: $\mathbf{h}_i = \text{NodeEncoder}(P_i)$.
  \vspace{-5pt}
  \item Startup embedding: $\mathbf{h}_{v_{target}} = \text{NodeEncoder}(\mathcal{S})$.
\end{itemize}
These embeddings form the input graph $\mathcal{G}_{\text{embed}}$ for VGAT to infer interaction edges:
\begin{equation}
\mathcal{E}_{\text{interact}} =  \text{VGAT}(\mathcal{G}_{\text{embed}}).
\end{equation}
These edges define the peer influence topology for the second decision round.

Agents revise their decisions by incorporating insights from interaction neighbors. Specifically, agent $A_i$ updates $D_i^{(0)}$ to $D_i^{(1)}$:
\begin{equation}
D_i^{(1)} = \text{LLM}\left(\left( D_i^{(0)}, \left\{ D_j^{(0)}, P_j \mid (A_i, A_j) \in \mathcal{E}_{\text{interact}} \right\} \right) \mid  (\mathcal{S}, P_i) \right).
\end{equation}

The overall investment confidence score $C_{\text{invest}}$ is defined as the proportion of agents opting to invest:
\begin{equation}
C_{\text{invest}} = \frac{1}{k} \sum_{i=1}^{k} \delta_i ,
\label{eq:conf-score}
\end{equation}
where $\delta_i = 1$ if $D_i^{(1)}$ is an investment decision, and 0 otherwise. This score aggregates individual assessments and collective peer influence.

\section{Experiments}
This section presents the main results; consistency analysis, prompt details, and explainability cases are in the Appendix.

\subsection{Datasets}
\label{sec:Datasets}


\noindent\textbf{PitchBook.} We compile a comprehensive global venture capital dataset from PitchBook\footnote{\url{https://pitchbook.com/}} covering the period from 2005 to November 2023. The dataset includes 263,729 startups and 1,014,157 individuals, with rich information on investment histories, demographic backgrounds, and detailed company profiles. For evaluation, we focus on 2,507 startups that received their initial funding between September 2021 and November 2022. The task is to determine whether a startup secures follow-on financing within one year of its initial funding round, resulting in 533 positive samples and 1,974 negative samples. To avoid look-ahead bias, all startup profiles are constructed exclusively using information available prior to the initial funding event. 

\noindent\textbf{Crunchbase.} Besides the private dataset, we used an open-source 2013 Crunchbase snapshot~\footnote{https://github.com/datahoarder/crunchbase-october-2013}. However, its age raises data leakage concerns; we report results in the Appendix 1.3.


\vspace{-5pt}
\subsection{Evaluation Metrics}

While standard classification metrics (e.g., F1 score) are reported, we prioritize Average Precision at K (AP@K) to align with the practical VC objective of identifying top-tier opportunities from a large candidate pool. In our proposed SimVC-CAS, to reflect real-world co-investment dynamics, a startup is considered positively predicted if at least one investor agent chooses to invest; rankings are then determined by the confidence score $C_{\text{invest}}$, mirroring how investment decisions aggregate in practice. Following prior work~\cite{sharchilev2018web,zhang2021scalable,10.1145/3763001}, we use Precision at K (P@K) to measure the success rate among the top-K ranked startups. To ensure temporal robustness, AP@K is computed by averaging P@K across monthly rolling windows. This metric directly quantifies the model's efffectiveness in shortlisting high-potential investments over time.



\vspace{-5pt}

\subsection{Implementation Details}
We employ GPT-3.5\footnote{\url{https://platform.openai.com/docs/models/gpt-3.5-turbo}} (knowledge cutoff: Sep 2021) on a test dataset collected post-September 2021 to \textbf{ensure strict temporal isolation and prevent data leakage}. For reproducibility, the temperature is fixed at 0. \textbf{The number of candidate investors $k$ is set to 10, which we empirically found to balance performance gains and computational efficiency (detailed analysis in Appendix 1.1)}. All reported results represent the average of five independent runs with random investor sampling.

\vspace{-5pt}

\subsection{Baselines}

To evaluate the effectiveness of our proposed method, we benchmark it against six diverse baselines: \textbf{Random}, which predicts based on historical success rates; \textbf{BERT Fusion}~\cite{maarouf2025fused}, which combines structured variables with BERT-encoded unstructured text for classification; \textbf{SHGMNN}~\cite{zhang2021scalable}, a heterogeneous graph neural network that learns representations end-to-end; \textbf{GST}~\cite{10.1145/3763001}, a spatiotemporal graph model that employs incremental graph self-attention to capture dynamic node representations; \textbf{GNN-RAG}~\cite{mavromatis2024gnn}, which synergizes GNN embeddings with RAG-retrieved shortest paths as context for LLMs; and \textbf{SSFF}~\cite{wang2025ssffinvestigatingllmpredictive}, a multi-agent framework that, unlike our interactive approach, analyzes dimensions in isolation before aggregating insights via a single centralized decision-maker.









\vspace{-3pt}

\begin{table}[!h]
\small 
\centering
\caption{Performance comparison on the test set}
\vspace{-12pt}
\begin{tabular}{lcccccc}
\toprule
Method & AP@10 & AP@30 & Precision & Recall & F1 \\\midrule
Random & $-$  & $-$ & 19.98 & 19.02 & 19.49 \\
BERT Fusion & 24.67 & 25.33 & 23.63  & 24.95  & 24.27 \\  
SHGMNN &25.41 & 26.22 & 20.51 & 82.37 & 32.97 \\
GST &25.71  &27.86  & 21.75 &\textbf{83.54}  &34.51 \\
GNN-RAG &27.53  &26.17  &22.81  &71.10 & 34.54 \\
SSFF & 30.02 & 27.26 &  23.23 & 69.41 & 34.81\\
$\textbf{\text{SimVC-CAS}}$  & \textbf{37.52}  &\textbf{30.83}  &\textbf{24.19}  &74.07  &\textbf{36.47} \\
\bottomrule
\end{tabular}
\vspace{-12pt} 
\label{tab:baseline_performance1}
\end{table}

\subsection{Experimental Results}

\textbf{The main experimental results \textbf{on the PitchBook dataset} are presented in Table~\ref{tab:baseline_performance1}.}
The proposed model significantly outperforms baselines across all evaluation metrics except recall, demonstrating its strong effectiveness in predicting the early-stage startup success. In particular, our SimVC-CAS achieves substantial gains on ranking-based metrics, with AP@10 improving by 25.0\%, and AP@30 by 13.1\% relative to the strongest baseline. \textbf{These results highlight SimVC-CAS's superior capability to prioritize and rank startups with high success potential.}

Further comparative analysis shows that graph-based models, including both traditional GNN approaches (e.g., SHGMNN and GST) and LLM RAG-enhanced variants like GNN-RAG, consistently underperform relative to SimVC-CAS. This underscores the importance of explicitly modeling potential investor relationships surrounding startups. Additionally, the notably weaker performance of the traditional multi-agent method SSFF further \textbf{validates the advantage of incorporating multi-decision-maker interaction dynamics over single-decision mechanisms.}

Furthermore, we find that while GNN methods (such as GST) achieve high recall (83.54\%) through aggressive prediction, their precision drops significantly (21.75\%). SimVC-CAS demonstrates overwhelming superiority in the most decision-critical AP@10 metric (37.52\% vs 25.71\%, a relative improvement of 46.1\%), with precision and overall F1 scores also significantly outperforming GNN methods.

\textbf{Results on the public Crunchbase dataset align with our main findings}, showing improvements of +6.82\% in AP@10, +2.66\% in AP@30, and +1.66\% in F1 over the strongest baseline (see Appendix 1.3 for details).

\begin{table}[!h]
\small 
\centering

\setlength{\tabcolsep}{2.5pt} 

\renewcommand{\arraystretch}{0.9}
\vspace{-4pt}
\caption{Ablation study and comparison of interaction methods}
\vspace{-10pt} 
\label{tab:Comparison of interaction methods}

\begin{tabular}{lcccc} 
\toprule
Method & AP@10 & AP@30 & F1 & \begin{tabular}{@{}c@{}}Decision\\Consistency\end{tabular} \\
\midrule

$w/o \ roleplay \ (single \ LLM)$ & 26.58 &  23.44 & 33.14 & - \\ 
${w/o \ interaction}$   & 35.43 & 28.04 & 35.62 & 0.68  
\\\midrule
$FullInteraction$ &32.14  &27.85  & 33.64 &  0.62 \\ 
$NetworkInteraction$ &36.43 & 29.28  & 35.87 & 0.67 \\ 
$GATInteraction$ &36.89  & 29.74 & 36.07 & 0.69 
\\\midrule
$\text{SimVC-CAS}$ & 37.52  &30.83  &36.47 &  0.71 \\
$\text{SimVC-CAS (4o-mini)}$ & 38.11 & 31.24 & 37.02 &  0.73 \\ 
$\text{SimVC-CAS (Claude 3.5)}$ & 38.27  & 31.79 & 37.39 &  0.74 \\ 
\bottomrule
\end{tabular}
\vspace{-12pt} 
\end{table}
\subsection{Ablation Study}


Table~\ref{tab:Comparison of interaction methods} shows ablation results. \textbf{$w/o \ roleplay$}, where a single LLM jointly processes the startup profile and all $k$ investor resumes without simulating individual personas, suffers the most significant performance drop. \textbf{This further validates that centralized reasoning cannot substitute for the collective intelligence of diverse agents.} Similarly, \textbf{$w/o \ interaction$} removes the interaction module and underperforms the full model, demonstrating the effectiveness of our interaction design.

We further compare different interaction strategies. $FullInteraction$, where all agents communicate indiscriminately, performs worse than the non-interaction baseline ($w/o \ interaction$) (F1: 33.64 vs 35.62), \textbf{suggesting that unstructured communication introduces detrimental noise.} In contrast, $NetworkInteraction$, which restricts interactions to historical co-investors, outperforms both (F1: 35.87), \textbf{validating the importance of leveraging established co-investment topologies.} Finally, while $GATInteraction$ improves upon $NetworkInteraction$, it still lags behind our full model (F1: 36.07 vs 36.47). \textbf{This confirms that our VGAT mechanism, by explicitly modeling the startup context, captures interaction patterns more effectively than standard GNNs.}

Finally, benchmarking with upgraded backbones (GPT-4o-mini\footnote{\url{https://platform.openai.com/docs/models/gpt-4o-mini}}, Claude 3.5 Sonnet\footnote{\url{https://docs.anthropic.com/en/docs/about-claude/models}}) yields consistent improvements, confirming the robustness of our framework across different LLMs. However, we acknowledge that \textbf{data leakage} in these newer models likely contributes to the gains.

\subsection{Consistency with Real Investors}
To evaluate whether our agents simulate real investor behavior, we introduce the \textbf{Decision Consistency} metric. For all investor-startup pairs, this metric measures the alignment between simulated and actual decisions, defined as the percentage of simulated decisions matching ground truth, for which the random baseline is 0.50.\footnote{We use a balanced dataset (all positive samples + equal random negatives) to calculate Decision Consistency. In raw VC data, ``Invest" decisions are rare, so consistency is dominated by ``Pass" decisions—allowing trivial ``always Pass" strategies to score high. Balancing forces the metric to equally weight the agent's ability to identify ``Invest" opportunities and correctly ``Pass".}
Despite potential noise where investors simply missed unseen opportunities rather than rejecting them, this metric validates that our agents capture specific preferences beyond chance.

As shown in Table~\ref{tab:Comparison of interaction methods}, 
$\text{SimVC-CAS}$ achieves \textbf{the highest Decision Consistency (0.71)}, 
outperforming $w/o \ interaction$ (0.68) and the random baseline (0.50). 
The failure of $FullInteraction$ (0.62) confirms that \textbf{our selective, 
graph-based interaction design is essential} for modeling real VC dynamics.


\subsection{Decision Benefit from Modeling Interactions}

To quantify the marginal contribution of the collective interaction mechanism, we introduce \textbf{Confidence Alignment Gain} ($\Delta C$), which captures whether interaction steers predictions closer to ground truth:
\vspace{-2pt}
\begin{equation}
\Delta C = (C_{\text{invest}}^{(1)} - C_{\text{invest}}^{(0)}) \cdot (2y - 1), \quad y \in \{0, 1\}, 
\end{equation}

where $C_{\text{invest}}^{(1)}$ and $C_{\text{invest}}^{(0)}$ are confidence scores after and before interaction, and $y$ is the ground-truth label. The term $(2y-1)$ ensures that $\Delta C > 0$ indicates improved alignment with ground truth—higher confidence for successful startups or lower confidence for failed ones.

\begin{figure}[h]
    \centering
    \vspace{-5pt}
    \includegraphics[width=1\linewidth]{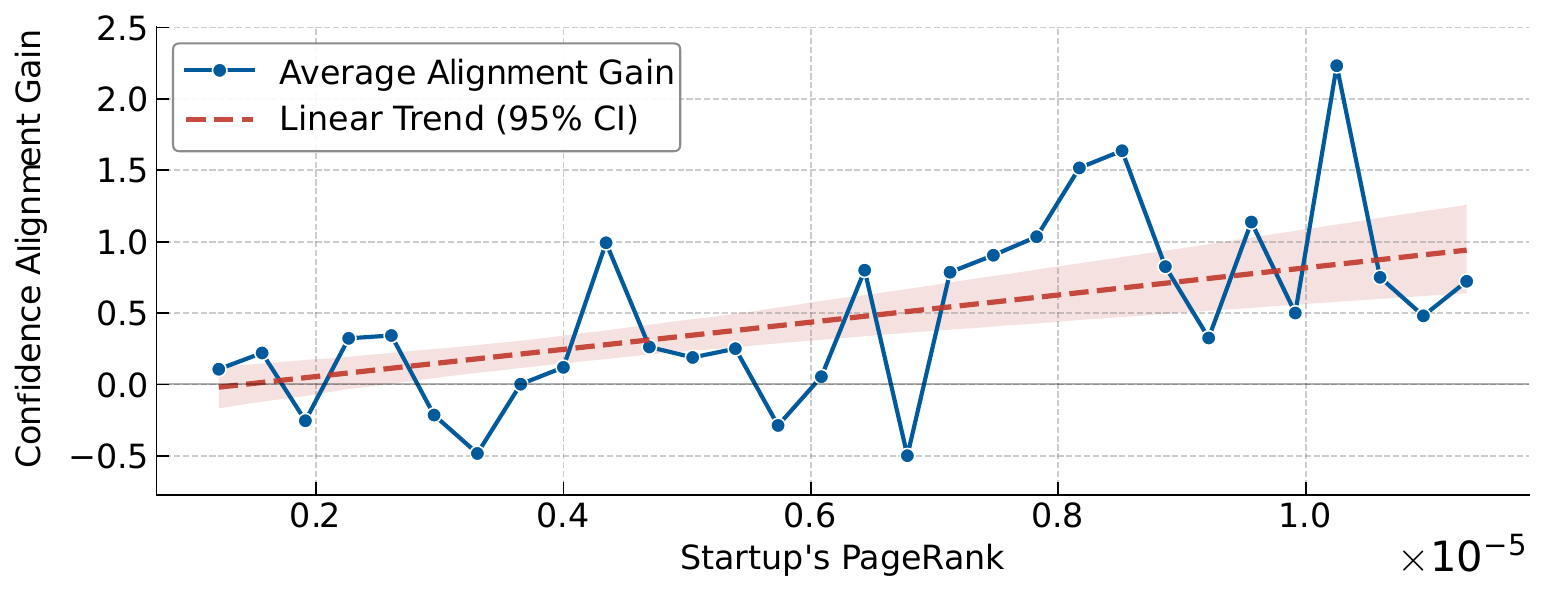}
        \vspace{-12pt}
    \caption{Impact of startup network centrality on collective interaction gain. }
    \label{fig:pagerank_gain}
     \vspace{-12pt}
\end{figure}

We investigate the correlation between this gain and startup network 
centrality (Figure~\ref{fig:pagerank_gain}). We calculate the PageRank 
score for each startup in the test set based on the investor-startup 
bipartite graph and compute the mean $\Delta C$ across PageRank bins. 
A linear regression with 95\% confidence interval reveals a clear trend: 
confidence alignment gain increases with PageRank. A higher PageRank 
indicates that the startup is embedded in a denser VC network, making 
it more critical to capture complex investor interaction dynamics. 
The trend validates that \textbf{for startups embedded in active investment networks, our interaction design is particularly effective.}

\subsection{Why Change Decisions After Interaction?}

\begin{figure}[h] 
    \vspace{-8pt}
    \centering
    \includegraphics[width=1\linewidth]{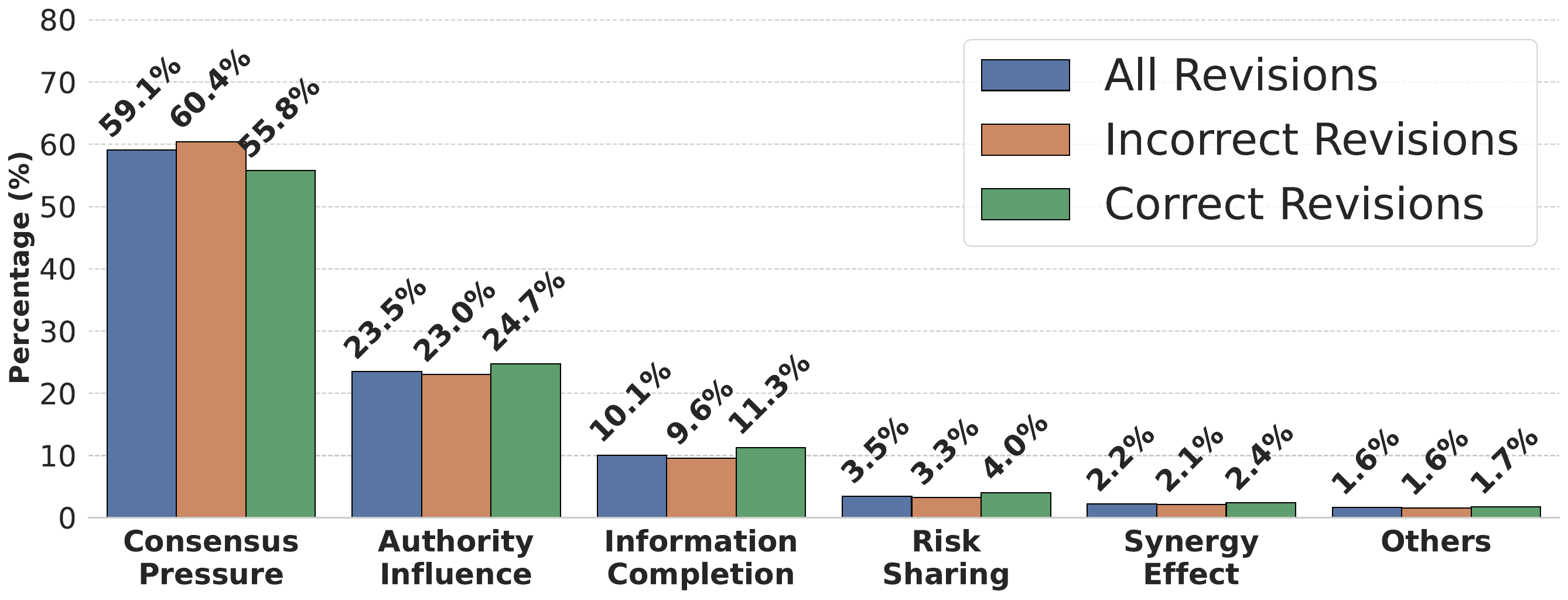} 
    \caption{Reasons for decision revisions across groups.}
    \label{fig:reason_dist}
        \vspace{-5pt}
\end{figure}

To understand what drives decision changes, we prompt agents who revised their decisions after interaction to select from predefined reasons. As shown in Figure~\ref{fig:reason_dist} (bars in blue), \textbf{Consensus Pressure (59.1\%), Authority Influence (23.5\%), and Information Completion (10.1\%)} emerge as the most frequently cited factors. We further distinguish between correct revisions (changes aligning with ground truth, in green) and incorrect revisions (orange). Notably, \textbf{correct revisions rely less on conforming to majority opinions} from surrounding agents (55.8\% vs 60.4\%), but \textbf{place more weight on deferring to established investors} (24.7\% vs 23.0\%) and \textbf{incorporating new factual details} through the network (11.3\% vs 9.6\%), suggesting that \textbf{effective decisions prioritize expert opinions and facts from the network over surrounding consensus}~\cite{sorensen2007smart,banerjee1992simple}. See Appendix 1.4 for classification details.


\section{Conclusion}
From the perspective of investor groups, this study proposes SimVC-CAS, 
a collective agent system designed to simulate the decision-making 
dynamics of venture capital. Unlike traditional approaches that overlook 
the investor collective effect, SimVC-CAS explicitly models the 
interactions among investors during the investment decision-making 
process. Our experiments demonstrate that this collective modeling 
approach achieves significant performance improvements and reveals how 
network environment shape decision quality, particularly for 
network-central startups. The SimVC-CAS framework is designed to be 
general and may extend to other complex group 
decision-making settings.


\clearpage

\bigskip

\bibliographystyle{named}
\bibliography{ijcai26}

\section{Appendix}

\subsection{Impact of Candidate Investor Scale ($k$)}

\begin{figure}[htbp]
    \centering
    \vspace{-5pt}
    \includegraphics[width=1\linewidth]{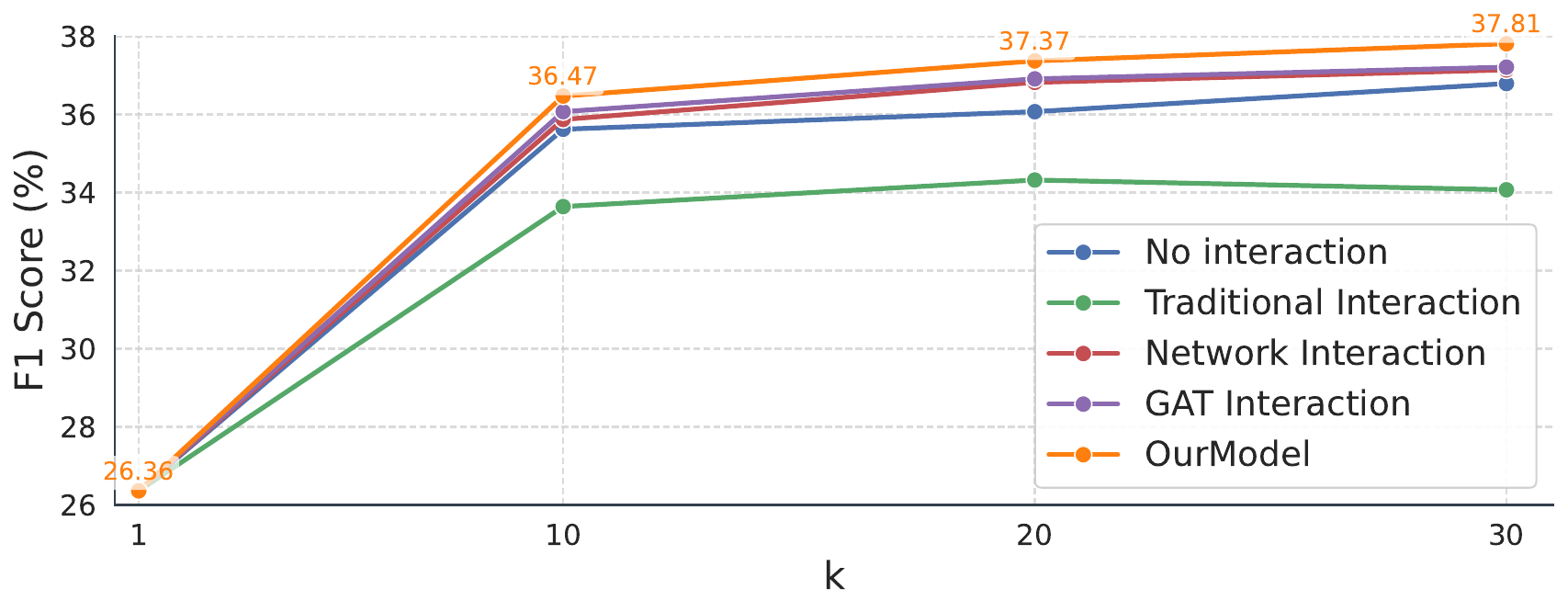}
    \vspace{-12pt}
    \caption{Performance comparison (F1 score) across varied interaction strategies with different $k$ values. }
    \label{fig:Comparison_k}
    \vspace{-15pt}
\end{figure}
To evaluate the impact of the number of candidate investors $k$ on our method, we conducted experiments with $k=1, 10, 20,$ and $30$. Meanwhile, we examined the performance differences across various interaction modes under different $k$ values. As shown in Figure~\ref{fig:Comparison_k}, which presents F1 score line graphs for each interaction mode across the tested $k$ values, all interaction modes - except for the traditional interaction - demonstrate improved F1 as $k$ increases. However, the rate of improvement diminishes significantly when $k > 10$. \textbf{Based on these observations, we consider $k=10$ to be a cost-effective choice: randomly selecting 10 investors around a startup provides a representative approximation of the broader investor group's decision-making. Given that real-world VC decisions involve substantial capital and complex evaluations, the computational overhead introduced by this model is minimal in comparison. Therefore, scaling up the number of candidate investors emerges as a practical and effective strategy for enhancing model performance.} 

Furthermore, it was found that when k=1, the F1 is lower than $w/o \ roleplay$ (26.36 vs 30.03). At this point, the model does not involve multi-role interaction and is essentially a few-shot learning with biases. \textbf{This indicates that relying solely on the experience of a single investor for few-shot learning will introduce significant decision biases, resulting in the model's performance being weaker than the zero-shot capability of the LLM.}

\subsection{Details of VGAT training}

To simulate the investor interaction pattern jointly determined by the characteristics of the target company, the attributes of individual investors and the topology of the co-investment network, we designed the VGAT model. In the method section of the main text, we have already elaborated on the details of VGAT. Here, we will introduce the training process of VGAT.  We train  VGAT  using the AdamW optimizer with a learning rate configuration of $ 1\mathrm{e}^{-3} $.  The contrast learning hyperparameter $\alpha$ set to 0.01.  All the aforementioned experiments are conducted on two RTX 3090 GPUs.  All the aforementioned experiments are conducted on two RTX 3090 GPUs.

\subsubsection{Introduction to Training Data}

Our training objective is to utilize supervised learning to enable the model to capture the interaction patterns among investors. The data records the historical investment information of each startup company. We use its future joint investment relationship with investors as a supervisory signal - this is a stricter indicator for measuring investor interaction. Although this standard is rather strict and may lose some potential interaction signals, the main experimental results in the text show that this method is effective.

To ensure no time overlap with the main evaluation task (based on companies that received their first round of financing from October 2021 to November 2022), we selected the data of companies that received their first round of financing from October 2016 to September 2021 for training and evaluation. The company used needs to meet the condition of obtaining the next round of financing within the next year, which provides the required data on future joint investment relationships. The dataset is divided as follows: Training set: Companies that received their first round of financing from October 2016 to September 2019; Validation set: Companies that received their first round of financing from October 2019 to March 2020; Test set: Companies that received their first round of financing from April 2020 to September 2020.

For each eligible startup company, in order to be consistent with the main experiment, we randomly selected $k$ = 10, 20, and 30 potential candidate investors respectively to construct three independent joint investment relationship prediction networks. In each network, if two investors connected by an edge do indeed jointly invest in the company within the next year, that edge is marked as a positive sample. Otherwise, it is marked as a negative sample.

Since candidate investors are randomly selected, the resulting co-investment graphs are naturally sparse, leading to extreme class imbalance in some instances. To ensure training stability and prevent the model from collapsing into a trivial "always-negative" predictor, we filter out networks with an extremely low signal-to-noise ratio (positive-to-negative edge ratio $< 0.05$). After this necessary preprocessing to handle sparsity, the final dataset comprises 1,992 training networks, 247 validation networks, and 222 test networks. Crucially, even after filtering, the average positive-to-negative ratio remains low at 0.2062, preserving the challenge of identifying sparse interaction signals while removing only the most uninformative noise

\subsubsection{Evaluation indicators and model performance}

To quantitatively evaluate our model, we assess its classification performance by computing the average Precision (P), average Recall (R), and average F1-score (F1) for graphs within each test set.

To highlight the rationality of the VGAT design through comparison, we adopted a two-layer GAT model as the baseline model for comparison. In this baseline model: 1. Virtual nodes are not introduced; 2. The embedding representation of company information is directly concatenated after the feature vectors of each investor node. 3. The characteristics of an edge are ultimately formed by the characteristics of the two associated investor nodes and the concatenate of the edge weights. In addition, the other structural designs and loss functions of the model are consistent with those of VGAT.
\begin{table}[htbp]
\small 
\centering
\begin{tabular}{lccccccc}
\toprule
Method & Precision & Recall & F1 \\\midrule
GAT &69.62 &77.76  & 73.47   \\
VGAT &\textbf{76.18}  &\textbf{82.70}  &\textbf{79.31}  \\

\bottomrule
\end{tabular}
\caption{The performance of GAT and VGAT}
\label{tab:VGAT_performance1}
\end{table}

As shown in Table \ref{tab:VGAT_performance1}, VGAT demonstrates significant advantages in the joint investment relationship prediction task. Compared with the baseline GAT model, VGAT improved by 6.56\%, 4.94\% and 5.84\% respectively in the three indicators of precision (76.18\% vs 69.62\%), recall (82.70\% vs 77.76\%) and F1 value (79.31\% vs 73.47\%). This consistency advantage validates the effectiveness of the virtual node design: by explicitly modeling the global impact of company characteristics on investment interaction patterns, VGAT more accurately captures the collaborative behaviors among investors.

\subsection{Crunchbase Public-Dataset Evaluation}
\label{app:crunchbase}

\paragraph{Dataset.}
To facilitate reproducibility, we use the publicly available Crunchbase 2013 snapshot\footnote{\url{https://github.com/mtwilliams/crunchbase-in-2013}} containing 196,553 companies and 226,708 individuals funded between 2005--2013. \textbf{Although smaller in scale, it preserves relational structures consistent with PitchBook.}

\paragraph{Task and labels.}
We align our task formulation with the main experiments: predicting whether a startup that receives seed/angel funding at time $t_0$ will successfully raise Series A capital within the subsequent year. The binary label is defined as $y=1$ if the next financing round occurs during the interval $[t_0, t_0{+}12\text{ months}]$, and $y=0$ otherwise.
\paragraph{Splits.}

Data is partitioned strictly by time to prevent temporal information leakage, resulting in an approximate train:val:test ratio of 7:1:2. The dataset yields a total of 4,375 labeled samples, with the training and validation periods strictly preceding the testing window. As discussed in the main text, given the vintage of this public dataset, we acknowledge that the LLMs employed in our framework may have been exposed to this historical data during pre-training, making implicit data leakage unavoidable.

\paragraph{Training and baselines.}
All model components, prompts, and evaluation protocols follow the main experiments. We compare SimVC-CAS against the same baselines under identical metric definitions.

\paragraph{Results.}
Table~\ref{tab:crunchbase_results} reports results on Crunchbase. 
SimVC-CAS achieves consistent improvements over the strongest baseline: \textbf{+6.82\%} AP@10, \textbf{+2.66\%} AP@30, and \textbf{+1.66\%} F1, matching the trends observed on PitchBook. \textbf{We also note that all models exhibit varying degrees of performance degradation on Crunchbase compared to PitchBook, likely due to the fact that this public snapshot contains less comprehensive information than the proprietary PitchBook dataset.}

\begin{table}[htbp]
\centering
\small
\setlength{\tabcolsep}{4pt}
\renewcommand{\arraystretch}{1.10}
\caption{Performance on public Crunchbase data.}
\begin{tabular}{lccccc} 
\toprule
\textbf{Method} & \textbf{AP@10} & \textbf{AP@30} & \textbf{Precision} & \textbf{Recall} & \textbf{F1} \\
\midrule
BERT Fusion & 25.57 & 25.26 & 23.54 & 48.10 & 31.61 \\
SHGMNN      & 26.38 & 26.13 & 18.85 & 64.82 & 29.12 \\
GST         & 26.54 & 26.32 & 19.64 & \textbf{68.45} & 30.48 \\
GNN-RAG     & 27.12 & 25.26 & 20.88 & 58.34 & 30.70 \\
SSFF        & 27.36 & 25.41 & 22.15 & 51.92 & 31.05 \\
\midrule
\textbf{SimVC-CAS} & \textbf{34.18} & \textbf{28.98} & \textbf{24.05} & 54.12 & \textbf{33.27} \\
\bottomrule
\end{tabular}
\label{tab:crunchbase_results}
\end{table}

\subsection{Taxonomy and Classification of Decision Revision}
\label{app:taxonomy}

\subsubsection{Task Definition}
To interpret the collective dynamics within SimVC-CAS, we analyze the rationale behind investors' decision updates. specifically, for every investor agent who changes their decision (e.g., from ``Pass" to ``Invest" or vice versa) after the interaction phase, we aim to categorize the primary driver of this revision. This analysis provides qualitative insights into how the simulated agents influence each other.

\subsubsection{Category Taxonomy}
We establish a taxonomy of six decision revision reasons, grounded in classic venture capital and behavioral finance literature:

\begin{enumerate}
    \item \textbf{Information Completion:} The investor revises their decision after acquiring critical overlooked information (e.g., market details, technical barriers) from peers. \textit{(Basis: Information Asymmetry \& Social Learning)}
    
    \item \textbf{Authority Influence:} The investor is swayed by the opposing judgment of a more senior or reputable investor (e.g., a lead investor). \textit{(Basis: Reputation Effect \& Status)}
    
    \item \textbf{Consensus Pressure:} The investor alters their confidence upon perceiving that the majority of peers hold a contrary opinion. \textit{(Basis: Herding Behavior)}
    
    \item \textbf{Synergy Effect:} The decision changes because the entry of specific investors is expected to bring complementary resources. \textit{(Basis: Value-Added Syndication)}
    
    \item \textbf{Risk Sharing:} The investor decides to invest after seeing others join, thereby reducing the perceived idiosyncratic risk. \textit{(Basis: Risk Diversification)}
    
    \item \textbf{Others:} Any rationale that does not fit into the above categories (e.g., purely random fluctuations or vague explanations).
\end{enumerate}

\subsubsection{Automated Classification with GPT-4o}
To scale this analysis across all test samples, we employ \textbf{GPT-4o} as an automated annotator. The model is provided with the investor's self-reported ``Revised Reasoning" and tasked with mapping it to one or more of the defined categories. The prompt used for this classification is presented in Figure~\ref{fig:reasoning_prompt}.

\begin{figure}[ht] 
    \centering
    \begin{tcolorbox}[colback=gray!10, colframe=black, title=Prompt for Decision Revision Classification]
    \small
    \textbf{Role:} You are a professional Venture Capital Analyst. \\
    \textbf{Task:} Analyze the reason why an investor changed their decision after communicating with peers. Read the [Decision Revision Explanation] and classify it into one or more of the following categories: \\
    
    \textbf{Categories:} \\
    A. [Information Completion]: Gained overlooked key info (market, tech, team) from peers. \\
    B. [Authority Influence]: Influenced by senior investors or lead investors. \\
    C. [Consensus Pressure]: Confidence changed due to majority opposing opinion. \\
    D. [Synergy Effect]: Expected resource complementarity from specific co-investors. \\
    E. [Risk Sharing]: Perceived risk reduced due to others joining. \\
    F. [Others]: Reasons not fitting above. \\
    
    \textbf{Input:} \\
    {}[Decision Revision Explanation]: \{Input\_Text\} \\ 
    
    \textbf{Output Requirement:} \\
    Output only the matched Category ID and Name (e.g., ``A. [Information Completion]; C. [Consensus Pressure]").
    \end{tcolorbox}
    \caption{The prompt used for classifying decision revision reasons.}
    \label{fig:reasoning_prompt}
\end{figure}

\subsection{Consistency Analysis}
Unlike embedding based models, large language models may produce different results over multiple runs. Therefore, we analyzed the consistency of the results generated by SimVC-CAS. We randomly select 20 startups and repeat the experiment 20 times.

\begin{figure}[htbp]
    \centering
    \includegraphics[width=0.9\linewidth]{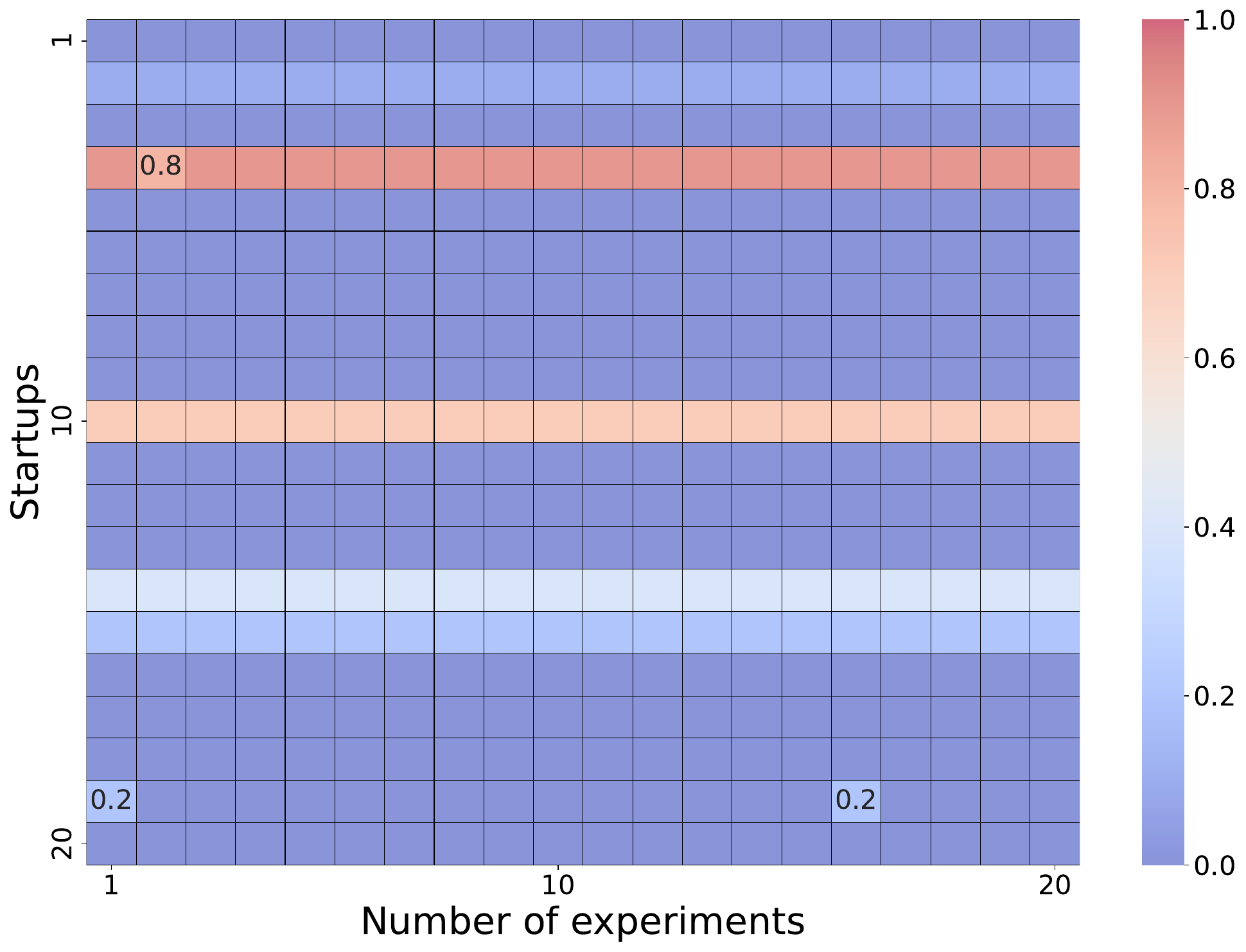}
    \caption{The consistency experiment results.}
    \label{fig:Consistency}
\end{figure}

The experimental results are shown in Figure ~\ref{fig:Consistency}. The horizontal axis represents different experimental results, the vertical axis represents each start-up company, and different colors reflect various scores output by the system. Overall, only a very small number of samples showed differences in results across different experiments, which fully demonstrates that our model has excellent consistency performance.

\subsection{Prompt Display}

\begin{figure}[htbp]
    \centering
    \includegraphics[width=1\linewidth]{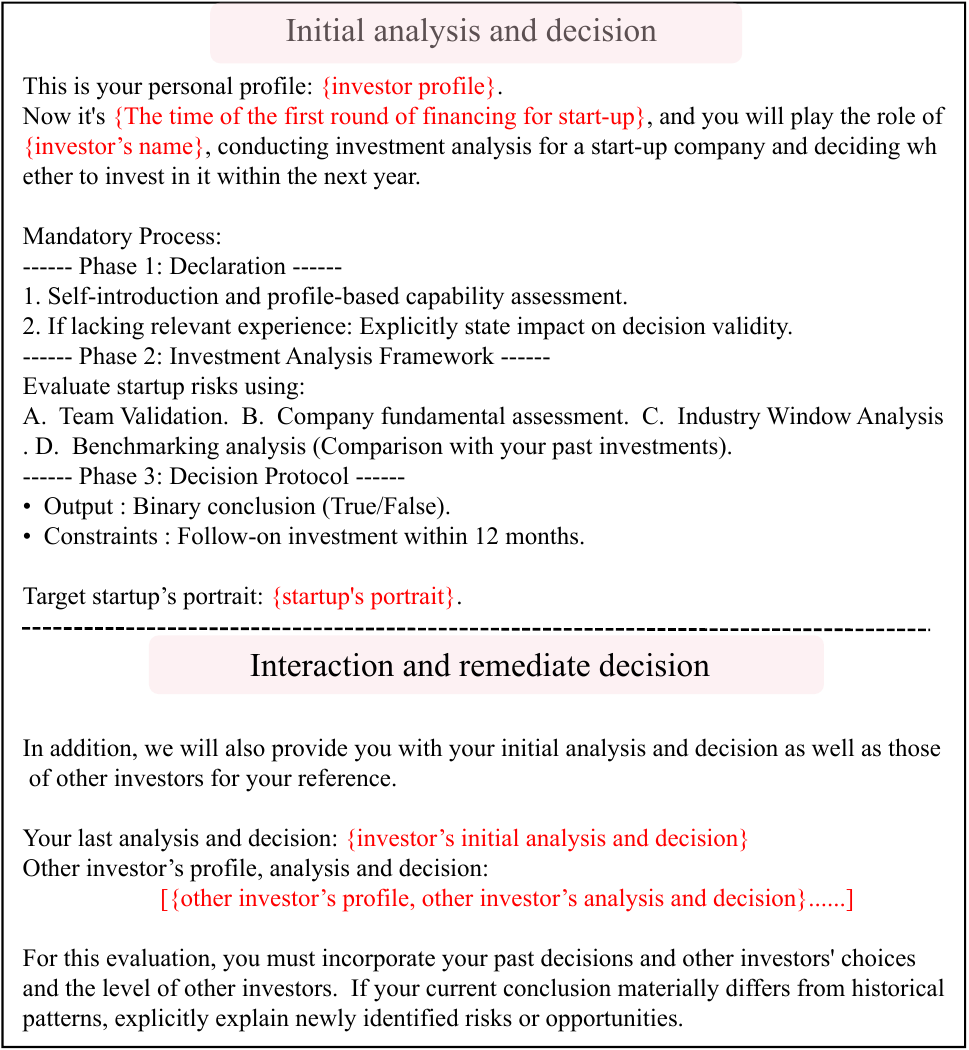}
    \caption{Prompt display}
    \label{fig:prompt}
\end{figure}

As shown in Figure \ref{fig:prompt}, we present the design ideas of the prompts used in the model for initial decision-making and interactive corrective decision-making.

\subsection{Explainability Display}
\begin{figure*}[htbp]
    \centering
    \includegraphics[width=1\linewidth]{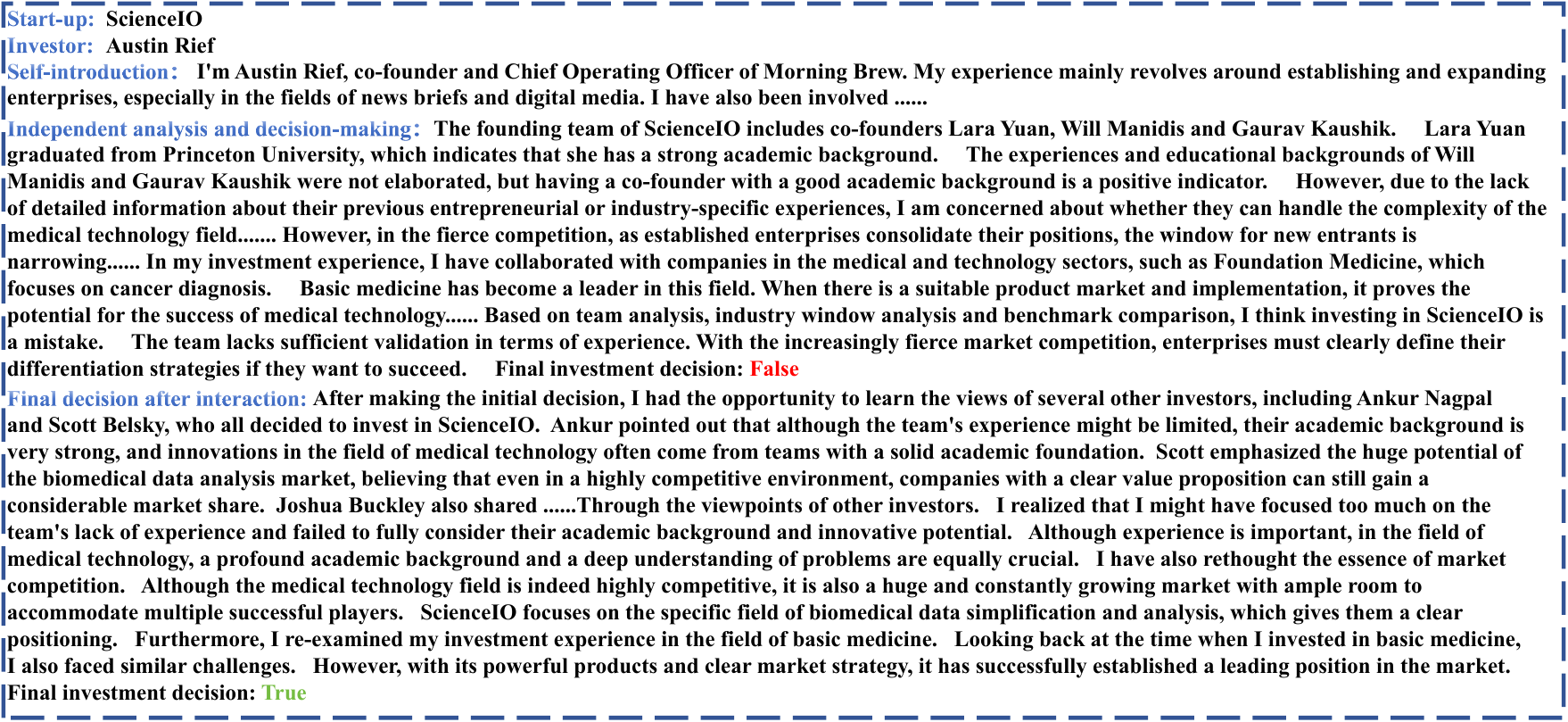}
    \caption{An example demonstrating how investors make investment decisions and ultimately revise their investment decisions through interaction with other investors.}
    \label{fig:example1}
\end{figure*}

\begin{figure*}[htbp]
    \centering
    \includegraphics[width=1\linewidth]{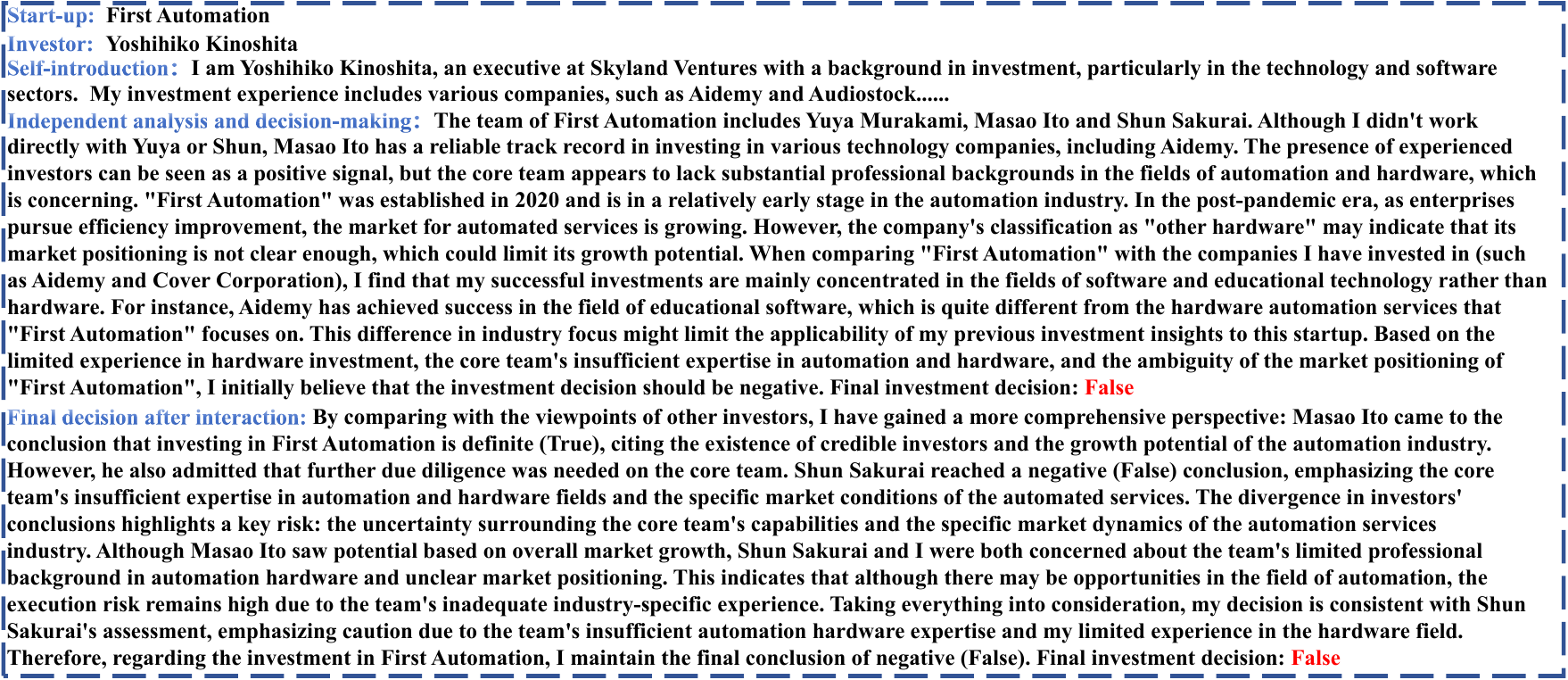}
    \caption{An example demonstrating how investors make investment decisions and ultimately revise their investment decisions through interaction with other investors.}
    \label{fig:example2}
\end{figure*}

As shown in the figure~\ref{fig:example1} and figure~\ref{fig:example2}, we present two example to illustrate the interpretability of the SimVC-CAS prediction results. In this example, the investor first introduced himself, then presented his independent analysis and investment decision, and finally, after interacting with other investors, the investor revised his decision.

\end{document}